\documentclass[10pt,journal,compsoc]{IEEEtran}
\ifCLASSOPTIONcompsoc
  \usepackage[nocompress]{cite}
\else
  \usepackage{cite}
\fi
\ifCLASSINFOpdf
\else
\fi
\usepackage{graphicx}
\usepackage{caption}
\usepackage{enumerate}
\usepackage{amsmath}
\usepackage{graphicx}
\usepackage{tabularx}
\usepackage{multirow, multicol}
\usepackage{subfigure}
\usepackage{algorithm}
\usepackage{amssymb}
\usepackage{bbm}
\usepackage{algpseudocode}
\usepackage{booktabs}

\begin{document}
%
\title{Structure-enhanced Contrastive Learning for Graph Clustering}
%
%
%
%

\author{Xunlian~Wu,
        Jingqi~Hu,
        Anqi~Zhang,
        Yining~Quan,
        Qiguang~Miao,~\IEEEmembership{Senior~Member,~IEEE}
        and~Peng~Gang~Sun
        
\IEEEcompsocitemizethanks{\IEEEcompsocthanksitem * Corresponding author 
\IEEEcompsocthanksitem Xunlian Wu, Jingqi Hu, Yining Quan, Qiguang Miao and Peng Gang Sun are with School of Computer Science and Technology Xidian University, Xi’an 710071, China.\protect\\
Email:\{xunlianwu@stu.xidian, jingqihu@stu.xidian, ynquan@xidian, qgmiao@xidian, psun@mail.xidian\}.edu.cn.
\IEEEcompsocthanksitem Anqi Zhang is with School of Computer and Artificial Intelligence, Henan Finance University, Zhengzhou, 450046, China.\protect\\
Email: zhanganqi@hafu.edu.cn}
\thanks{Manuscript received April 19, 2005; revised August 26, 2015.}}

%
%

\markboth{Journal of \LaTeX\ Class Files,~Vol.~14, No.~8, August~2015}%
{Shell \MakeLowercase{\textit{et al.}}: Bare Demo of IEEEtran.cls for Computer Society Journals}
%



\IEEEtitleabstractindextext{%
\begin{abstract}
Graph clustering is a crucial task in network analysis with widespread applications, focusing on partitioning nodes into distinct groups with stronger intra-group connections than inter-group ones. Recently, contrastive learning has achieved significant progress in graph clustering. However, most methods suffer from the following issues:  1) an over-reliance on meticulously designed data augmentation strategies, which can undermine the potential of contrastive learning. 2) overlooking cluster-oriented structural information, particularly the higher-order cluster(community) structure information, which could unveil the mesoscopic cluster structure information of the network. In this study, Structure-enhanced Contrastive Learning (SECL) is introduced to addresses these issues by leveraging inherent network structures. SECL utilizes a cross-view contrastive learning mechanism to enhance node embeddings without elaborate data augmentations, a structural contrastive learning module for ensuring structural consistency, and a modularity maximization strategy for harnessing clustering-oriented information. This comprehensive approach results in robust node representations that greatly enhance clustering performance. Extensive experiments on six datasets confirm SECL's superiority over current state-of-the-art methods, indicating a substantial improvement in the domain of graph clustering.
\end{abstract}

}

\maketitle

\IEEEdisplaynontitleabstractindextext

%
\IEEEpeerreviewmaketitle

\IEEEraisesectionheading{\section{Introduction}\label{sec:introduction}}

%
%
%
%
\IEEEPARstart{G}{raph} clustering, also known as community detection, is an essential aspect of complex network analysis, drawing significant interest for its capability to segment networks into clusters with densely interconnected nodes. This segmentation yields insights into the network's structure and is crucial for applications in social networking, biology, and recommendations \cite{coscia2011classification,ying2018graph,xing2022comprehensive}. However, when faced with complex graph data that includes high-dimensional attribute information, it presents a challenge in effectively processing the graph data. Traditional methods, such as those based on sequences \cite{perozzi2014deepwalk,grover2016node2vec}, non-negative matrix factorization \cite{wang2017community,zhu2021community}, and spectral clustering \cite{ng2001spectral,hu2020community}, only utilize the structural information of the graph. Methods based on attribute clustering \cite{krishna1999genetic,xie2016unsupervised} only use attribute information. This exclusive focus on one type of data, either structure or attributes, can reduce the performance of clustering.

The advent of Graph Neural Networks (GNNs) enables the simultaneous processing of attribute and structural information. With their advanced graph representation learning capabilities, several GNN-based methods \cite{krishna1999genetic,pan2018adversarially, zhang2019attributed} have been proposed for graph clustering. End-to-end graph clustering methods have been developed that directly yield clustering outcomes to learn embeddings with a clustering tendency \cite{wang2019attributed,zhou2023community,wu2024deep}. Additionally, to address the issue of over-smoothing, SDCN \cite{bo2020structural} combines GAE with standard autoencoders, forming a unified framework to capture higher-order structural information. Although these methods are effective for graph clustering, their dependence on accurately initialized cluster centers may result in a tedious and inefficient pre-training process.

To address the aforementioned issues, contrastive learning, as a self-supervised learning method, substitutes the cluster update optimization loss with a contrastive loss, thus mitigating the manual trial-and-error problem. Inspired by this, a series of graph clustering methods grounded in contrastive learning have been proposed \cite{you2020graph,cui2020adaptive,zhu2021graph,hassani2020contrastive,zhao2021graph,liu2022deep,liu2023simple}, achieving noteworthy performance. However, we observe that these methods exhibit two primary issues: 1) The success of contrastive learning is heavily contingent upon carefully selected graph data augmentations. Inappropriate graph data augmentations, including random attribute perturbation or random edge dropping, may induce semantic drift, compromising the integrity of the learned embeddings. 2) They overlook clustering-oriented information, particularly the higher-order cluster(community) structure information, which could unveil the mesoscopic cluster structure information of the network. The mesoscopic cluster structure pertains to the organization of the graph at an intermediate level, bridging the local structure captured by individual node connections and the global structure of the entire graph.

To address these issues, we propose a novel method termed Structure-enhanced Contrastive Learning for Graph Clustering (SECL). Eschewing complex data augmentation techniques for graph perspectives, we leverage the inherent network structure, obviating the necessity for carefully designed data augmentation. We construct two perspectives through the generation of attribute and structural encodings that respectively capture the graph's attribute and structural information. A cross-view contrastive loss is introduced to learn more discriminative node embeddings by contrasting graph representations under different views. To reinforce structural information, a structural contrastive loss is employed, aiming to align the cross-view similarity matrix with the self-looped adjacency matrix. Finally, to advance the model's assimilation of clustering-oriented information, we devised a modularity maximization loss function. This loss function motivates the model to optimize embeddings so as to reflect the cluster (community) structure within the graph. In summary, the principal contributions of this study are as follows:

\begin{itemize}
	\item We propose a novel Structure-Enhanced Contrastive Learning (SECL) approach that aims to improve the performance of graph clustering tasks without the need for pre-training and carefully designed  data augmentation. The core of this method lies in directly extracting valuable information from the graph structure itself and utilizing this information to enhance the model's learning process.
	\item We combine contrastive learning with modularity optimization for the task of graph clustering. This approach aims to enhance the model's ability to recognize community structures in graphs through the framework of contrastive learning. By this integration, the model can learn representations by leveraging both the similarities between nodes and the overall community structure of the graph.
	\item Extensive experimental results on six different domain datasets demonstrate the superiority of SECL over state-of-the-art graph clustering methods.
\end{itemize}
\section{Related work}
In this section, we provide a concise overview of recent advancements in two interrelated areas: graph clustering and contrastive learning.

\subsection{Graph Clustering}
Graph clustering aims to partition the nodes of a graph into disjoint sets by learning a graph embedding. This methodology is essential for uncovering the underlying structure of complex networks. The growing importance of graph clustering has garnered significant attention from the research community, leading to the development of numerous algorithms tailored for this purpose \cite{liu2022survey,xing2022comprehensive}. 

Learning high-quality embeddings for graph clustering is crucial; traditional graph embedding methods \cite{perozzi2014deepwalk,grover2016node2vec,lee1999learning,ye2018deep,zhu2020simple} often overlook the attribute or structural information of networks. However, thanks to the development of Graph Convolutional Neural Networks (GCNs), these tools incorporate both the attribute and structural information of networks to learn node embeddings. Kipf et al. \cite{kipf2016variational} introduced the GAE and VGAE models, employing autoencoders to reconstruct adjacency matrices and facilitate graph embedding. Building on this, MGAE \cite{wang2017mgae} refines the GAE approach by enhancing the embedding process. In a related vein, ARGA and ARVGA \cite{pan2018adversarially} leverage these graph autoencoders to effectively extract embedding information. Furthering the development of these models, AGC \cite{zhang2019attributed} utilizes higher-order graph convolutions to identify global clustering patterns. Addressing the common issue of over-smoothing in such models, SDCN \cite{bo2020structural} integrates the principles of GAE with standard autoencoders, creating a cohesive framework that captures higher-order structural data. The DAEGC \cite{wang2019attributed} algorithm takes advantage of an attention network to assign weights to nodes, optimizing for both reconstruction loss and KL-divergence-based clustering loss. In a similar context, CDBNE \cite{zhou2023community} uses a graph attention mechanism to process topology and node attributes, aiming to adeptly identify community structures within networks by maximizing modularity. DDGAE \cite{wu2024deep} integrates high-order modularity information and attribute information as distinct views and learns the latent node representations through the reconstruction of topology, attribute, and modularity information.

While the previously mentioned graph clustering techniques are proficient, their dependency on accurately initialized cluster centers for optimal performance can result in a laborious and inefficient trial-and-error pre-training process \cite{pan2018adversarially,wang2019attributed,bo2020structural,cheng2021multi}. In contrast, the application of contrastive learning alters this dynamic, as it substitutes the traditional cluster update optimization loss function with a contrastive loss. This innovative approach negates the necessity for manual trial-and-error pre-training. 
\subsection{Contrastive Learning}
Contrastive learning, as a self-supervised learning approach, focuses on learning distinctive features of data by maximizing the similarity between positive pairs and increasing the dissimilarity between negative pairs, which facilitates better clustering performance without the need for extensive pre-training \cite{liu2023simple,hassani2020contrastive,zhao2021graph}. SimCLR \cite{chen2020simple} constitutes a significant advancement in representation learning, aligning augmented versions of the same data instances to achieve maximum concordance. Extending beyond, GraphCL \cite{you2020graph} introduces four novel graph-specific augmentations that enhance unsupervised graph representation learning. GCA \cite{zhu2021graph} refines this method with an adaptive augmentation technique that adjusts to the unique structure and attributes of each graph. MVGRL \cite{hassani2020contrastive} leverages diffusion processes to generate diverse graph views, improving representation learning quality. SCAGC \cite{xia2021self} alters the graph topology by randomly adding or removing edges. GDCL \cite{zhao2021graph} combines graph diffusion with an innovative use of signal smoothness to guide its contrastive learning, while DCRN \cite{liu2022deep} employs a siamese network structure focusing on the nuanced hierarchical features in graph node representations. Despite their effectiveness, these approaches rely on dependable data augmentation for positive pair identification. Shifting towards more sophisticated strategies, CCGC \cite{yang2023cluster} and SCGC \cite{liu2023simple} produce dual augmented vertex views via distinct siamese encoders, representing a departure from conventional augmentation techniques. NCAGC \cite{wang2023neighborhood} pioneers a method that forgoes explicit graph data augmentation, instead utilizing a node's neighboring information to form contrastive pairs, thus leveraging the graph's inherent structure. CGC \cite{xie2023contrastive} represents a paradigm shift, utilizing the graph's connectivity to propel a contrastive learning framework, obviating the need for external augmentation and concentrating on innate inter-node connections.
\begin{figure*}[!h]
	\centering
	\includegraphics[width=0.99\linewidth]{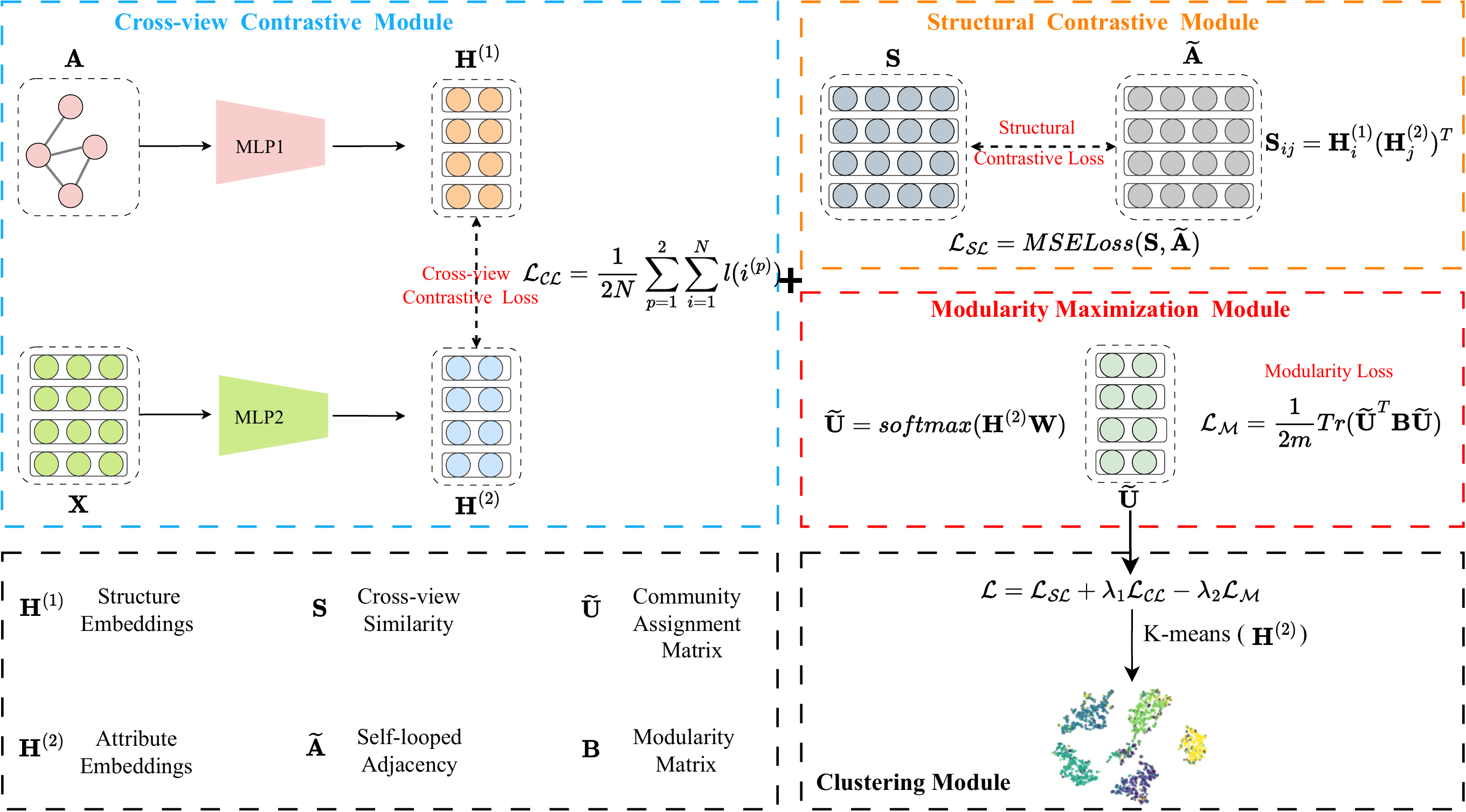}\\
	\caption{Overview of SECL applied to graph clustering. In the cross-view contrastive module, structure and attributes are first embedded into the latent space using the structure encoder $\text{MLP}^{(1)}$ and attribute encoder $\text{MLP}^{(2)}$, thereby bypassing the need for complex data augmentation. Subsequently, the similarity between the attribute and structure embeddings is computed to derive the cross-view contrastive loss. Next, within the structure contrastive loss module, consistency of structure information is ensured by aligning the similarity matrix with the neighboring structure information. Then, a modularity maximization module is employed to capture cluster-oriented information. Finally, we jointly optimize the cost functions of three modules using the Adam optimizer. Post-optimization, the graph clustering results are obtained by applying K-means to the attribute embeddings $\textbf{H}^{(2)}$.}\label{model}
\end{figure*}
\section{Proposed Approach}
In this section, we present a novel Structure-enhanced contrastive learning method (SECL) for graph clustering. The overall architecture of our model is illustrated in Figure \ref{model}. The model comprises three main modules: the Cross-view Contrastive module, which enhances node embeddings without elaborate data augmentations; the Structural Contrastive Module, which ensures the consistency of structure information by aligning the similarity matrix with the neighboring structure information and the Modularity Maximization module is employed to capture cluster-oriented information. In the subsequent sections, we will elaborate on all modules and the objective function. 
\begin{table}[!t]
	\centering
	\caption{Notations and Descriptions.}
	\small
	\scalebox{1.0}{
		\begin{tabular}{ll}
			\toprule
			\textbf{Notations}                                        & \textbf{ Descriptions}                                \\ \midrule
			$\textbf{X}\in \mathbb{R}^{N\times d}$  & Attribute matrix  
			\\
			$\widehat{\textbf{X}}\in \mathbb{R}^{N\times d}$  & Graph filtered attribute matrix
			\\
			$\textbf{A}\in \mathbb{R}^{N\times N}$  & Original adjacency matrix   
			\\
			$\widetilde{\textbf{A}}\in \mathbb{R}^{N\times N}$  & Adjacency matrix with self-loop
			\\
			$\widehat{\textbf{A}}\in \mathbb{R}^{N\times N}$  & Symmetric normalized adjacency matrix
			\\
			$\textbf{L}\in \mathbb{R}^{N\times N}$  & Graph Laplacian matrix
			\\
			$\widehat{\textbf{L}}\in \mathbb{R}^{N\times N}$  & Symmetric normalized Laplacian matrix
			\\
			$\textbf{H}^{(1)} \in \mathbb{R}^{N\times d'}$   & Structure embeddings       
			\\
			$\textbf{H}^{(2)} \in \mathbb{R}^{N\times d'}$  & Attribute embeddings 
			\\
			$\textbf{S} \in \mathbb{R}^{N\times N}$ & Cross-view similarity
			\\
			$\textbf{U}\in \mathbb{R}^{N\times C}$ & Community assignment matrix
			\\
			$\textbf{B} \in \mathbb{R}^{N \times N}$ & Modularity matrix
			\\
			\bottomrule
	\end{tabular}}
	\label{NOTATION_TABLE} 
\end{table}

\subsection{Problem Formulation}
Given a graph $\mathcal G=(\mathcal V,\mathcal E,\textbf{A},\textbf{X})$, let $\mathcal V=\{v_1,v_2,\ldots,v_n\}$ represent a set of $N$ nodes categorized into $C$ communities, and $\mathcal E=\{(v_i,v_j)|v_i,v_j\in \mathcal V\}$ constitutes a set of $m$ edges. $\textbf{X}\in \mathbb{R}^{N\times d}$ denotes the attribute matrix. The original adjacency matrix is defined as:
\begin{equation}
	\textbf A=\left\{\begin{array}{ll}
		\textbf A_{i,j}=1,&\textrm{$(v_i,v_j)$ $\in$ $\mathcal E$}\\
		\textbf A_{i,j}=0,&\textrm{$(v_i,v_j)$ $\notin$ $\mathcal E$},
	\end{array} \right.
\end{equation} 
The degree matrix is defined as:
\begin{equation}
	\textbf{D}=diag(d_1,d_2,\ldots,d_n)\in \mathbb R^{N\times N},
\end{equation}
Where $d_i=\sum_{(v_i,v_j)\in \mathcal E}\textbf{A}_{ij}$.
The graph Laplacian matrix is expressed as:
\begin{equation}
	\textbf {L}=\textbf{D}-\textbf{A},
\end{equation}
Employing the renormalization trick $\widetilde{\textbf{A}}=\textbf{A}+\textbf{I}$, which introduces a self-loop to each node, the symmetric normalized adjacency matrix is defined as:
\begin{equation}
	\widehat{\textbf{A}}=\widetilde{\textbf{D}}^{-\frac{1}{2}}\widetilde{\textbf{A}}\widetilde{\textbf{D}}^{-\frac{1}{2}},
\end{equation}
Thus, the graph Laplacian matrix is rewritten as:
\begin{equation}
	\widehat{\textbf{L}}= \textbf{I}-\widehat{\textbf{A}},
\end{equation}
Table \ref{NOTATION_TABLE} provides a summary of the primary notations used throughout the paper, along with their corresponding meanings.

Deep graph clustering aims to partition nodes into $C$ disjoint groups in an unsupervised manner, where nodes within the same group exhibit denser connections than with nodes outside their group.

\subsection{Encoders}
In this section, we introduce two distinct types of encoders aimed at embedding nodes into a latent space: the structure encoder ($\text{MLP}^{(1)}$) and the attribute encoder ($\text{MLP}^{(2)}$), which are responsible for encoding the structure information and the attribute information of the samples, respectively.

To capture the structural information of nodes more effectively, we propose a dedicated structure encoder. The design of the structure encoder $\text{MLP}^{(1)}$ is detailed as follows:
\begin{equation} 
	\textbf{H}^{(1)} = \text{MLP}^{(1)}(\textbf{A}),
	\label{MLP1}
\end{equation}
Where $\textbf{H}^{(1)}$ denotes structure embeddings of the nodes.

Prior to attribute encoding, we employ a commonly used Laplacian filter \cite{cui2020adaptive,liu2023hard} to aggregate and assimilate neighbor information, effectively filtering out high-frequency noise. The process is as follows:
\begin{equation}
	\widehat{\textbf{X}} = (\prod_{i=1}^{r}(\textbf{I} - \widehat{\textbf{L}}))\textbf{X} = (\textbf{I} - \widehat{\textbf{L}})^r\textbf{X},
	\label{SMOOTH}
\end{equation}
Where $\widehat{\textbf{X}}$ denotes the smoothed attribute matrix, $\widehat{\textbf{L}}$ represents the symmetric normalized Laplacian matrix, and $\textbf{I} - \widetilde{\textbf{L}}$ constitutes the graph Laplacian filter, with $r$ signifying the number of layers of graph Laplacian filters applied. Subsequently, we encode the smoothed attribute matrix $\widehat{\textbf{X}}$ using $\text{MLP}^{(1)}$ as follows:
\begin{equation} 
	\textbf{H}^{(2)} = \text{MLP}^{(2)}(\widehat{\textbf{X}}),
	\label{MLP2}
\end{equation}
Where $\textbf{H}^{(2)}$ denotes attribute embeddings of the nodes. Finally, we normalize $\textbf{H}^{(1)}$ and $\textbf{H}^{(2)}$ as follows:
\begin{equation}
	\textbf{H}^{(1)} = \frac{\textbf{H}^{(1)}}{||\textbf{H}^{(1)}||_2};
	\textbf{H}^{(2)} = \frac{\textbf{H}^{(2)}}{||\textbf{H}^{(2)}||_2},
\end{equation}
Where $||\cdot ||_2$ denotes $\ell^{2}$-norm. $\textbf{H}^{(1)}$ and $\textbf{H}^{(2)}$ represent the normalized structure embeddings and attribute embeddings, respectively, each reflecting a distinct view. The classical Mean-Square Error is used to train the two encoders. 
 
\subsection{Cross-view Contrastive Learning}
In this section, we introduce a cross-view contrastive module that utilizes contrastive learning methods to 'pull closer' the node representation of one view to its corresponding node representation of another view (positive) , while 'pushing it away' from the representations of other nodes(negative). This strategic approach thereby enhances the discrimination of representations. The similarity amongst node pairs is quantified as:
\begin{equation}
	\textbf S_{ij}^{(p,q)} = sim(\textbf{H}^{(p)}_{i},\textbf{H}^{(q)}_{j})/\tau,
	\label{pq}
\end{equation}
The similarity function, $sim(\textbf{H}_i,\textbf{H}_j)$, is defined as the transpose of the dot product $\textbf{H}_i\textbf{H}_j^T$, where $\tau$ denotes the temperature parameter in the similarity calculation. This parameter is instrumental in scaling the similarity scores and thus affects the separation of positive and negative sample pairs during the learning process. $p,q\in \{1,2\}$, if $p=1$, then $q=2$; conversely if $q=1$, then $p=2$. For a given node $i$ in $\textbf{H}^{(1)}$, this node can form a positive pair with the corresponding node $i$ in $\textbf{H}^{(2)}$, while the remaining $N-1$ nodes in $\textbf{H}^{(2)}$ constitute negative pairs. Then, the cross-view contrastive loss is formulated as:
\begin{equation}
	\ell (i^{(p)})=-log\frac{\exp(\textbf S_{ii}^{(p,q)})}{\sum_{k=1}^{N}\exp(\textbf S_{ik}^{(p,q)})},
\end{equation}
Where $i^{(p)}$ represents node $i$ in view $p$. Considering all nodes in graph $\mathcal G$ across all views, the proposed framework systematically calculates the cross-view contrastive loss as follows:
\begin{equation} 
	\mathcal{L_{CL}}=\frac{1}{2N}\sum_{p=1}^2\sum_{i=1}^N l(i^{(p)}),
	\label{CL}
\end{equation} 

This cross-view contrastive loss converges the embeddings of the same node from both the attribute and structural perspectives, while diverging the embeddings of distinct nodes. This approach thus augments the discriminative capability of the node representations. It is important to emphasize that different nodes within the same perspective are not considered negative samples. The aim is to capitalize on the distinctions between the two perspectives (attribute space and structural space). Employing the same node as a positive sample from different perspectives serves to fully harness the consistency across these perspectives, thereby enriching the node's representation by integrating diverse information sources.
\subsection{Structural Contrastive Learning}
In this section, to augment structural information, a strategy akin to SCGC \cite{liu2023simple} was adopted. To preserve the coherence of the cross-view structure information, Mean Squared Error ($MSE$) is utilized to force the cross-view similarity equal to an adjacency matrix with self-loops ($\widetilde{\textbf{A}}$). Similarly to the prior computation of $ \textbf{S}_{ij}^{(1,2)} $, the cross-view similarity matrix $\textbf{S} \in \mathbb{R}^{N \times N}$ is derived between $ \textbf{H}^{(1)} $ and $ \textbf{H}^{(2)} $. This matrix plays a pivotal role in aligning the node embeddings from different views and facilitates the integration of their respective attributes and structures.
\begin{equation}
	\textbf S_{ij} = \textbf{H}^{(1)}_{i}(\textbf{H}^{(2)}_{j})^T, \forall i,j\in [1,N],
	\label{S}
\end{equation}
Where $\textbf{S}_{ij}$ represents the cross-view similarity between node $i$ from the first view and node $j$ from the second view. Then, the cross-view node similarity matrix $\textbf{S} \in \mathbb{R}^{N \times N}$ is constrained to match the self-looped adjacency matrix $\widetilde{\textbf{A}} \in \mathbb{R}^{N \times N}$, ensuring that the similarity measurements align with the inherent graph structure and thus reinforcing the consistency of node representations across different views.
\begin{equation}
	\begin{aligned}
		\mathcal{L_{SL}}&=MSELoss(\textbf{S},\widetilde{\textbf{A}})\\ &=\frac{1}{N^2}\sum_i^N\sum_j^N (\textbf{S}_{ij}-\widetilde{\textbf{A}}_{ij})^2 \\ &= \frac{1}{N^2}(\sum_i^N\sum_j^N\mathbbm{1}_{ij}^{1}(\textbf{S}_{ij}-1)^2+\sum_i^N\sum_j^N\mathbbm{1}_{ij}^{0}\textbf{S}_{ij}^2),
	\end{aligned}
	\label{SL}
\end{equation}
Where $\mathbbm{1}_{ij}^{1}$ represents the case when $\widetilde{\textbf{A}}_{ij}=1$, and $\mathbbm{1}_{ij}^{0}$ represents the case when $\widetilde{\textbf{A}}_{ij}=0$. Our objective is to minimize the Mean Squared Error Loss ($\mathcal{L_{SL}}$), ensuring that each constituent component remains minimal to achieve a low overall loss. In the first scenario, where nodes $i$ and $j$ are neighbors, an increase in $\textbf{S}_{ij}$ is associated with a reduction in the loss. Conversely, in the second scenario, non-neighboring nodes $i$ and $j$ exhibit a decrease in $\textbf{S}_{ij}$, which contributes to a lower loss. We define cross-view neighbors of a given node as positive samples and non-neighboring nodes as negative samples. This neighbor-contrastive loss approach promotes convergence among neighboring nodes from different views while diverging non-neighbors, thus maintaining cross-view structural consistency and enhancing clustering performance.

\subsection{Modularity maximization}
In this section, we utilize modularity maximization to refine the learned node embeddings, preserving the inherent community structure of the network. To the best of our knowledge, this study represents the inaugural integration of modularity optimization with contrastive learning for the acquisition of node embeddings. Modularity, a concept introduced by Newman \cite{newman2006modularity}, quantifies the robustness of community structures and is delineated as:
\begin{equation}\label{modularity}
	Q=\frac{1}{2m}  Tr(\textbf U^T\textbf B\textbf U)
\end{equation}
Where $\textbf{B} \in \mathbb{R}^{N \times N}$ signifies the modularity matrix, and $\textbf B_{ij} = \textbf A_{ij} - \frac{k_i k_j}{2m}$ encapsulates the modularity relationship between node $i$ and node $j$. $\textbf A_{ij}$ refers to the elements of the adjacency matrix, $k_i$ to the degree of node $i$, $N$ to the total number of nodes, and $m$ to the total number of edges within the graph, respectively. $Tr(\cdot)$ denotes the trace of a matrix. The matrix $\textbf{U} \in \mathbb{R}^{N \times C}$ characterizes the cluster (community) memberships of nodes and is detailed as:
\begin{equation}
	\textbf U_{ik}=\left\{ \begin{array}{ll}
		1,&\textrm{$i$ $\in$ $C_k$}\\
		0,&\textrm{otherwise}
	\end{array} \right.
\end{equation}
Where $C$ denotes the number of communities, $C_k$ the $k$-$th$ community. Although maximizing modularity is an NP-hard problem, various methods exist to optimize $Q$. In this work, we employ the relaxation $Tr(\textbf{U}^\mathsf{T}\textbf{U}) = N$ to simplify the concept of modularity. We normalize the cluster (community) assignment matrix $\textbf{U}$ as follows:
\begin{equation}
	\widetilde{\textbf U}_{i k}=\frac{\sqrt{N}\cdot\sqrt{\textbf {U}_{i k}}}{\sum_{i}^N\sum_{k}^C\sqrt{\textbf {U}_{i k}}},
\end{equation}
Where $\widetilde{\textbf U}$ is the normalized community assignment matrix, with each row corresponding to a node's compact representation. Modularity serves as the metric for refining network representation learning, which guarantees that the resultant node embeddings encompass essential network properties and maintain intrinsic community structures, substantiating the feasibility of embedding optimization via modularity maximization \cite{yang2016modularity,zhou2023community}.
\begin{algorithm}[t]
	\small
	\caption{Structure-enhanced Contrastive Learning}
	\label{ALGORITHM}
	\flushleft{\textbf{Input}: Graph $\mathcal{G}=\{\textbf{A},\textbf{X}\}$; Number of clusters $C$; Iterations $L$; Graph filter times $r$; hyper-parameter $\tau$, $\lambda_1$, $\lambda_2$}\\
	\flushleft{\textbf{Output}: Clustering results $\textbf{R}$} 
	\begin{algorithmic}[1]
		\State Obtain the smoothed attributes matrix $\widehat{\textbf{X}}$ in Eq. \eqref{SMOOTH}
		\For{$i=1$ to $L$}
		\State Encode $\textbf{A}$ with $\text{MLP}^{(1)}$ in Eq. \eqref{MLP1}
		\State Encode $\widehat{\textbf{X}}$ with $\text{MLP}^{(2)}$ in Eq. \eqref{MLP2}
		\State Calculate cross-view similarity matrix $\textbf{S}^{(p,q)}$ and $\textbf{S}$ by Eq. \eqref{pq} and \eqref{S}
		\State Calculate cross-view contrastive loss $\mathcal{L_{CL}}$ by Eq. \eqref{CL}
		\State Calculate structural contrastive loss $\mathcal{L_{SL}}$ by Eq. \eqref{SL}
		\State Calculate modularity maximization loss $\mathcal{L_{M}}$ by Eq. \eqref{M}
		\State Update the whole framework by minimizing the overall objective loss $\mathcal{L}$ with Adam optimizer in Eq. \eqref{loss}
		\EndFor
		\State $\textbf{R}=$ K-means($\textbf{H}^{(2)}$, $C$).
		\State \textbf{return} $\textbf{R}$
	\end{algorithmic}
\end{algorithm}

Due to the fact that the number of clusters (communities) is typically much smaller than the number of nodes, especially in large-scale networks, utilizing the number of communities as the dimension for node embeddings can result in a loss of rich semantic information. To overcome this limitation, we employ a learnable fully connected layer $\textbf W\in \mathbb R^{d'\times C}$ \cite{zhou2023community}. Ultimately, the modularity maximization loss function is defined as follows:
\begin{equation}
	\mathcal{L_M}=\frac{1}{2m}  Tr(\widetilde{\textbf U}^T\textbf B\widetilde{\textbf U}),
	\label{M}
\end{equation}
Where $\widetilde{\textbf U}=softmax(\textbf{H}^{(2)}\textbf W)$. $Softmax$ is an operation that inherently includes normalization.

\subsection{Objective Function}
The overall optimized objective of SECL contains cross-view contrastive loss $\mathcal{L_{CL}}$, structural contrastive loss $\mathcal{L_{SL}}$, and modularity maximization loss $\mathcal{L_M}$.
\begin{equation}
	\mathcal{L}=\mathcal{L_{SL}}+\lambda_1\mathcal{L_{CL}}-\lambda_2\mathcal{L_M}.
	\label{loss}
\end{equation}
Where $\lambda_1$ and $\lambda_2$ are two trade-off hyper-parameter. Our goal is to minimize $\mathcal{L}$ during training. After training is completed, the learned attribute embeddings $\textbf{H}^{(2)}$ undergoes K-means clustering to yield the definitive clustering results.

The main process of SECL method is outlined in Algorithm \ref{ALGORITHM}.
\begin{table}[h]
	\centering
	\caption{Statistical Information of Six Datasets.}
	\small
	\scalebox{0.95}{
		\begin{tabular}{@{}cccccc@{}}
			\toprule
			\textbf{Dataset} & \textbf{Sample} & \textbf{Edge} & \textbf{Dimension} & \textbf{Class} \\ \midrule
			CORA  & 2708    & 5429  & 1433       & 7       \\
			CITESEER   & 3327    & 4732  & 3703      & 6       \\
			AMAP& 7650   & 119081 & 745       & 8       \\
			BAT   & 131    & 1038 & 81      & 4       \\
			EAT   & 399    & 5994  & 203       & 4       \\
			UAT & 1190   & 13599 & 239       & 4       \\ \bottomrule
	\end{tabular}}
	\label{DATASET_INFO} 
\end{table}
\begin{table*}[]
	\centering
	\setlength{\abovecaptionskip}{2pt}%
	\setlength{\abovecaptionskip}{8pt}%
	\caption{Experimental Results for Graph Clustering Task on Six Datasets. The Best is Highlighted with \textbf{Bold} and the runner-up is Highlighted with \underline{Underline}.}
	\label{EXP}
	\setlength{\tabcolsep}{1.28mm}{\resizebox{\linewidth}{!}{
			\begin{tabular}{@{}c|c|cccccccccccc|c@{}}
				\hline
				\toprule
				\textbf{Dataset}                    & \textbf{Metric} & \textbf{K-Means} & \textbf{DEC} & \textbf{S$^2$GC} & \textbf{GAE} & \textbf{DAEGC} & \textbf{ARGA} & \textbf{SDCN} & \textbf{DFCN}                    & \textbf{AGE} & \textbf{MVGRL} & \textbf{SCAGC} & \textbf{SCGC}                     &                       \textbf{SECL} \\ \midrule
				& ACC             &33.80±2.71   & 46.50±0.26       & 69.28±3.70  & 43.38±2.11   & 70.43±0.36    & 71.04±0.25   & 35.60±2.83    & 36.33±0.49                        & 73.50±1.83    & 70.47±3.70        & 60.89±1.21                         & \underline{73.88±0.88}   &\textbf{74.79±0.82} \\
				& NMI             &14.98±3.43& 23.54±0.34      & 54.32±1.92  & 28.78±2.97   & 52.89±0.69    & 51.06±0.52   & 14.28±1.91    & 19.36±0.87                       &\textbf{57.58±1.42}    & 55.57±1.54         & 39.72±0.72                        & 56.10±0.72   &\underline{56.88±0.91} \\
				& ARI             &08.60±1.95& 15.13±0.42       & 46.27±4.01  & 16.43±1.65   & 49.63±0.43    & 47.71±0.33  & 07.78±3.24    & 04.67±2.10                        & 50.60±2.14    & 48.70±3.94        & 30.95±1.42                         & \underline{51.79±1.59}   &\textbf{52.84±0.67} \\
				\multirow{-4}{*}{\textbf{CORA}} & F1            &30.26±4.46  & 39.23±0.17       & 64.70±5.53   & 33.48±3.05   & 68.27±0.57    & 69.27±0.39   & 24.37±1.04    & 26.16±0.50                        & 69.68±1.59    & 67.15±1.86        & 59.13±1.85                         &  \underline{70.81±1.96}    &\textbf{73.29±1.59}\\ \midrule
				& ACC             &39.32±3.17& 55.89±0.20       & 68.97±0.34  & 61.35±0.80   & 64.54±1.39    & 61.07±0.49   & 65.96±0.31    & 69.50±0.20                        & 69.73±0.24    & 62.83±1.59        & 61.16±0.72                         & \underline{71.02±0.77}   &\textbf{71.34±0.60} \\
				& NMI            &16.94±3.22 & 28.34±0.30     &42.81±0.20  & 34.63±0.65  & 36.41±0.86   & 34.40±0.71    & 38.71±0.32   & 43.90±0.20    & 44.93±0.53                        & 40.69±0.93    & 32.83±1.19                                 & \underline{45.25±0.45}   &\textbf{46.18±0.91} \\
				& ARI            &13.43±3.02 & 28.12±0.36     &44.42±0.32  & 33.55±1.18  & 37.78±1.24   & 34.32±0.70    & 40.17±0.43   & 45.50±0.30    & 45.31±0.41                       & 34.18±1.73    & 31.17±0.23                                  & \underline{46.29±1.13}   &\textbf{46.70±0.96} \\
				\multirow{-4}{*}{\textbf{CITESEER}}     & F1              &36.08±3.53& 52.62±0.17    &64.49±0.27   & 57.36±0.82  & 62.20±1.32   & 58.23±0.31    & 63.62±0.24   & 64.30±0.20    & 64.45±0.27                       & 59.54±2.17    & 56.82±0.43                                & \underline{64.80±1.01}   &\textbf{65.23±0.62} \\ \midrule
				& ACC             &27.22±0.76& 47.22±0.08       & 60.23±0.19  & 71.57±2.48   & 75.96±0.23    & 69.28±2.30   & 53.44±0.81    & 76.82±0.23                        & 75.98±0.68    & 41.07±3.12        & 75.25±0.10                         & \underline{77.48±0.37}    &\textbf{77.76±0.52}\\
				& NMI             &13.23±1.33& 37.35±0.05       & 60.37±0.15  & 62.13±2.79   & 65.25±0.45    & 58.36±2.76   & 44.85±0.83    & 66.23±1.21                        & 65.38±0.61    & 30.28±3.94        &  67.18±0.13                        &  \underline{67.67±0.88}   &\textbf{67.67±0.59} \\
				& ARI            &05.50±0.44 & 18.59±0.04       & 35.99±0.47  & 48.82±4.57   & 58.12±0.24    & 44.18±4.41   & 31.21±1.23    & 58.28±0.74                        & 55.89±1.34    & 18.77±2.34        & 56.86±0.23                        & \underline{58.48±0.72}    &\textbf{58.64±1.23}\\
				\multirow{-4}{*}{\textbf{AMAP}}   & F1          &23.96±0.51    & 46.71±0.12       & 52.79±0.01  & 68.08±1.76   & 69.87±0.54    & 64.30±1.95   & 50.66±1.49    & 71.25±0.31 & 71.74±0.93    & 32.88±5.50        & \textbf{72.77±0.16}                                             & 72.22±0.97   &\underline{72.29±0.59}\\ \midrule
				& ACC             &40.23±1.19& 42.09±2.21       & 36.11±2.16  & 53.59±2.04   & 52.67±0.00    & 67.86±0.80   & 53.05±4.63    & 55.73±0.06                        & 56.68±0.76    & 37.56±0.32        & 57.25±1.65                         & \underline{77.97±0.99}  &\textbf{78.40±0.69} \\
				& NMI            &26.92±2.39 & 14.10±1.99       & 13.74±1.60  & 30.59±2.06   & 21.43±0.35    & 49.09±0.54   & 25.74±5.71    & 48.77±0.51                        & 36.04±1.54    & 29.33±0.70        & 22.18±0.31                         & \underline{52.91±0.68}   &\textbf{53.74±0.68} \\
				& ARI             &09.52±1.42& 07.99±1.21        & 4.00±1.98  & 24.15±1.70   & 18.18±0.29    & 42.02±1.21   & 21.04±4.97    & 37.76±0.23                        & 26.59±1.83    & 13.45±0.03       & 27.29±1.53                        & \underline{50.64±1.85}  &\textbf{51.52±1.82} \\
				\multirow{-4}{*}{\textbf{BAT}}     & F1           &34.45±2.10   & 42.63±2.35       & 29.74±2.76  & 50.83±3.23   & 52.23±0.03    & 67.02±1.15   & 46.45±5.90    & 50.90±0.12                        & 55.07±0.80    & 29.64±0.49       & 52.53±0.54                        &  \underline{78.03±0.96}   &\textbf{78.47±0.57} \\ \midrule
				& ACC             &32.23±0.56& 36.47±1.60       & 32.41±0.45  & 44.61±2.10   & 36.89±0.15    & 52.13±0.00   & 39.07±1.51    & 49.37±0.19                        & 47.26±0.32    & 32.88±0.71        & 44.61±1.57                         &  \underline{57.94±0.42}    &\textbf{58.00±0.20}    \\
				& NMI             &11.02±1.21& 04.96±1.74       & 4.65±0.21  & 15.60±2.30   & 05.57±0.06    & 22.48±1.21   & 08.83±2.54    & 32.90±0.41                        & 23.74±0.90    & 11.72±1.08         & 07.32±1.97                        &  \underline{33.91±0.49}    &\textbf{33.98±0.27}    \\
				& ARI             &02.20±0.40& 03.60±1.87        &  1.53±0.04  & 13.40±1.26   & 05.03±0.08    & 17.29±0.50   &  06.31±1.95    & 23.25±0.18                        & 16.57±0.46    & 04.68±1.30         & 11.33±1.47                         &  \textbf{27.51±0.59}   &\underline{27.25±0.25} \\
				\multirow{-4}{*}{\textbf{EAT}}    & F1            &23.49±0.92  & 34.84±1.28       & 26.49±0.66  & 43.08±3.26   & 34.72±0.16    & 52.75±0.07   & 33.42±3.10    & 42.95±0.04                        & 45.54±0.40           & 25.35±0.75    & 44.14±0.24                         &  \underline{57.96±0.46}    &\textbf{58.15±0.25}\\ \midrule
				& ACC             &42.47±0.15& 45.61±1.84       & 36.74±0.81  &  48.97±1.52   &  52.29±0.49    & 49.31±0.15   & 52.25±1.91    & 33.61±0.09                        & 52.37±0.42    & 44.16±1.38        & 50.75±0.64                         &  \underline{56.58±1.62}  &\textbf{58.28±1.87}    \\
				& NMI             &22.39±0.69& 16.63±2.39       & 8.04±0.18  & 20.69±0.98   & 21.33±0.44    & 25.44±0.31   & 21.61±1.26    & 26.49±0.41                        & 23.64±0.66    & 21.53±0.94         & 23.60±1.78                        &  \underline{28.07±0.71}  &\textbf{28.96±1.33}    \\
				& ARI             &15.71±0.76& 13.14±1.97        & 5.12±0.27  & 18.33±1.79   & 20.50±0.51    & 16.57±0.31   & 21.63±1.49    & 11.87±0.23                        & 20.39±0.70    & 17.12±1.46         & 23.33±0.32                         &  \underline{24.80±1.85}   &\textbf{25.39±2.34} \\
				\multirow{-4}{*}{\textbf{UAT}}    & F1            &36.12±0.22  & 44.22±1.51       & 29.50±1.57  & 47.95±1.52   & 50.33±0.64    & 50.26±0.16   & 45.59±3.54    & 25.79±0.29                        & 50.15±0.73           & 39.44±2.19    & 47.07±0.73                         &  \underline{55.52±0.87}    &\textbf{56.66±1.03}\\
				\bottomrule \hline
			\end{tabular}
	}}
\end{table*}

\section{Experiments}
In this section, we conduct our experiments to showcase the effectiveness of our SECL model. 

\subsection{Datasets}
To validate the efficacy of our proposed methodology, datasets were curated from various domains \footnote{https://github.com/yueliu1999/Awesome-Deep-Graph-Clustering/tree/main/dataset}, including \textbf{CORA}, \textbf{CITESEER}, \textbf{AMAP}, \textbf{BAT}, \textbf{EAT}, and \textbf{UAT}. CORA and CITESEER are citation networks, where nodes represent academic papers and edges correspond to citations between them. The AMAP dataset originates from Amazon's collaborative purchasing graph, where nodes are products and edges indicate joint purchase frequencies. Meanwhile, the BAT, EAT and UAT datasets document passenger movements through airports over specific time intervals. Detailed information on the datasets is presented in the Table 1.
\subsection{Baselines}
To validate the superiority of our SECL, we have selected 12 algorithms for comparative experiments. This selection includes the classic clustering algorithm $\textbf{K-Means}$ \cite{krishna1999genetic}, which performs clustering directly based on the original attributes; the deep clustering algorithm $\textbf{DEC}$ \cite{xie2016unsupervised}, which performs clustering after learning node embeddings via an auto-encoder; and the spectral clustering algorithm $\textbf{S}^2\textbf{GC}$ \cite{zhu2020simple}, which balances low- and high-pass filters through the aggregation of K-hop neighbors. In addition, the selection also includes classic deep graph clustering algorithms like $\textbf{GAE}$ \cite{kipf2016variational}, $\textbf{DAEGC}$ \cite{wang2019attributed}, $\textbf{ARGA}$ \cite{pan2018adversarially}, $\textbf{SDCN}$ \cite{bo2020structural}, and $\textbf{DFCN}$ \cite{liu2022deep}, which cluster by learning node embeddings via graph auto-encoding. Moreover, included are the most advanced contrastive learning-based deep graph clustering algorithms like $\textbf{AGE}$ \cite{cui2020adaptive}, $\textbf{MVGRL}$ \cite{hassani2020contrastive}, $\textbf{SCAGC}$ \cite{xia2021self}, and $\textbf{SCGC}$ \cite{liu2023simple}.
\subsection{Experimental Setup}
The method realized through PyTorch, is implemented on an Intel(R) Core(TM) i9-10980XE CPU @ 3.00GHz, 128G of RAM, and NVIDIA GeForce RTX 3090 GPU and Ubuntu 18.04.6 LTS.

\textbf{Training procedure.}
The SECL algorithm is run for 400 epochs to achieve convergence, minimizing the loss function $\mathcal{L}$ with Adam \cite{kingma2014adam}. Subsequently, K-means clustering is applied to the resulting attribute embeddings $\textbf{H}^{(2)}$ to produce the final results. 

\textbf{Parameters setting.}
Our method utilizes the Adam \cite{kingma2014adam} optimizer for parameter learning, with the learning rate set to 1e-3 for the CORA/BAT/UAT datasets, 5e-2 for EAT, and 5e-5 for CITESEER/AMAP. Furthermore, the graph filters are configured with 2 for CITESEER, 3 for CORA/BAT/UAT, and 5 for AMAP/EAT. The temperature coefficient $\tau$ is assigned values of 0.1 for CORA/AMAP/BAT/UAT, 0.8 for CITESEER, and 1.0 for EAT. In the proposed method, for CORA/UAT/EAT, structure encoder $\text{MLP}^{(1)}$ and attribute encoder $\text{MLP}^{(2)}$ each comprise a single 500-dimensional embedding layer, and for CITESEER/AMAP/BAT, attribute encoder $\text{MLP}^{(2)}$ and structure encoder $\text{MLP}^{(1)}$ features two embedding layers with dimensions 1024 and 500.

\textbf{Evaluation criteria}
To assess the performance of graph clustering methods, we select four widely used metrics: Accuracy ($\textbf{ACC}$), Normalized Mutual Information ($\textbf{NMI}$), Adjusted Rand Index ($\textbf{ARI}$), and the F1-score ($\textbf{F1}$) \cite{wu2024deep,liu2023simple,wang2019attributed}. For each metric, a higher score indicates a more effective clustering.
\begin{figure*}[!t]
	\centering
	\subfigure[CORA]{
		\includegraphics[width=0.31\linewidth]{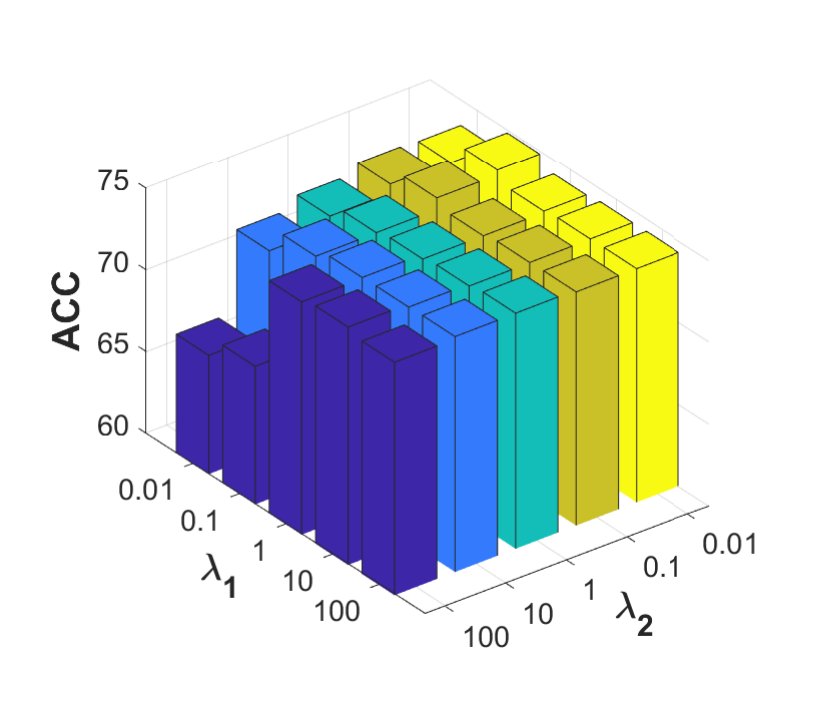}
	}
	\subfigure[CITESEER]{
		\includegraphics[width=0.31\linewidth]{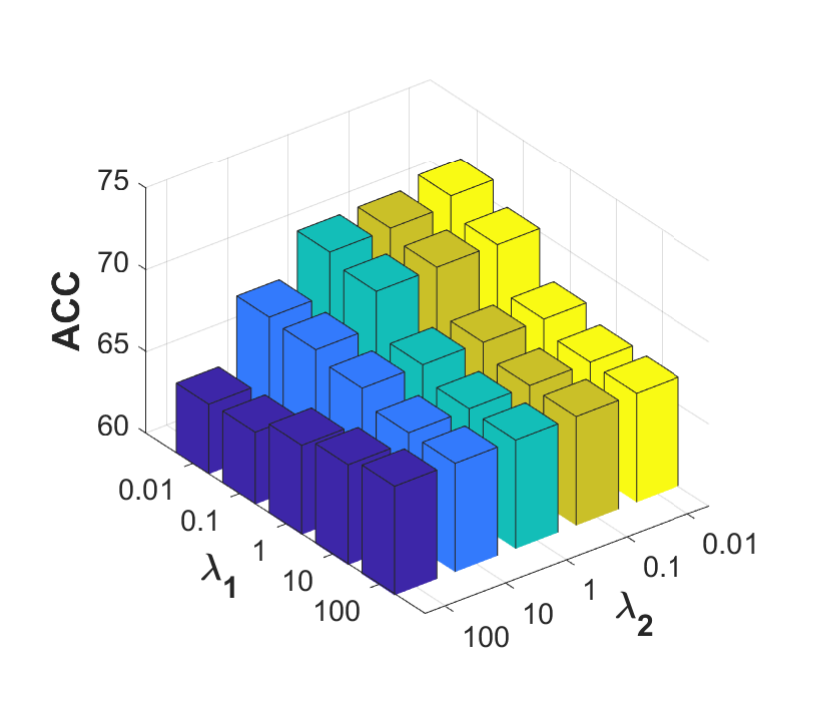}
	}
	\subfigure[AMAP]{
		\includegraphics[width=0.31\linewidth]{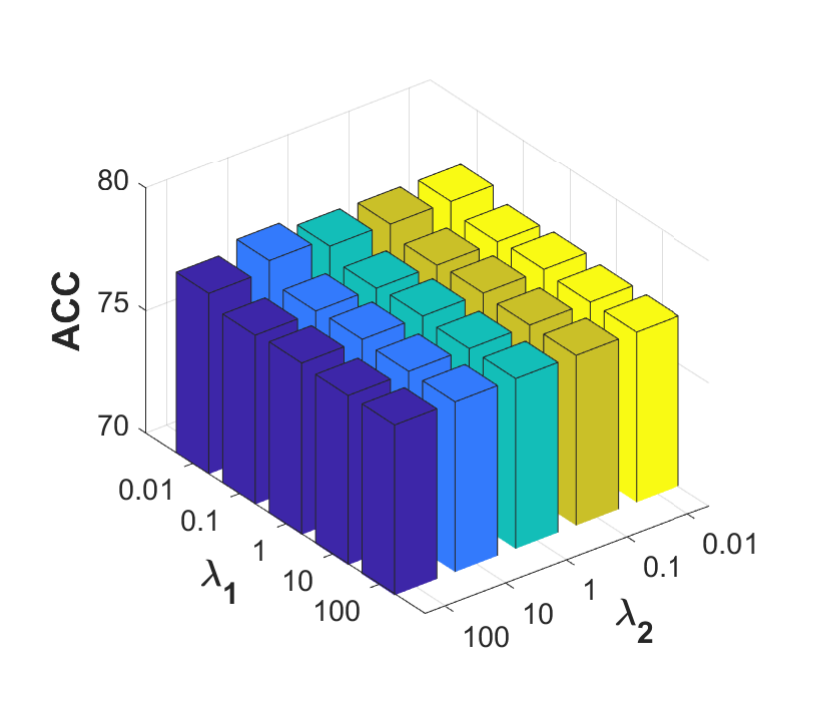}
	}
	\subfigure[BAT]{
		\includegraphics[width=0.31\linewidth]{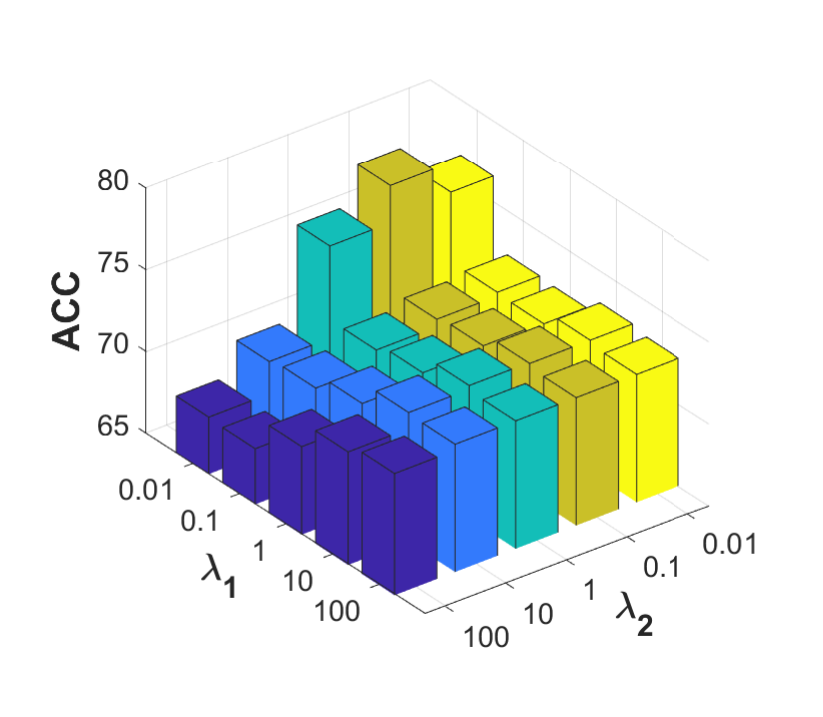}
	}
	\subfigure[EAT]{
		\includegraphics[width=0.31\linewidth]{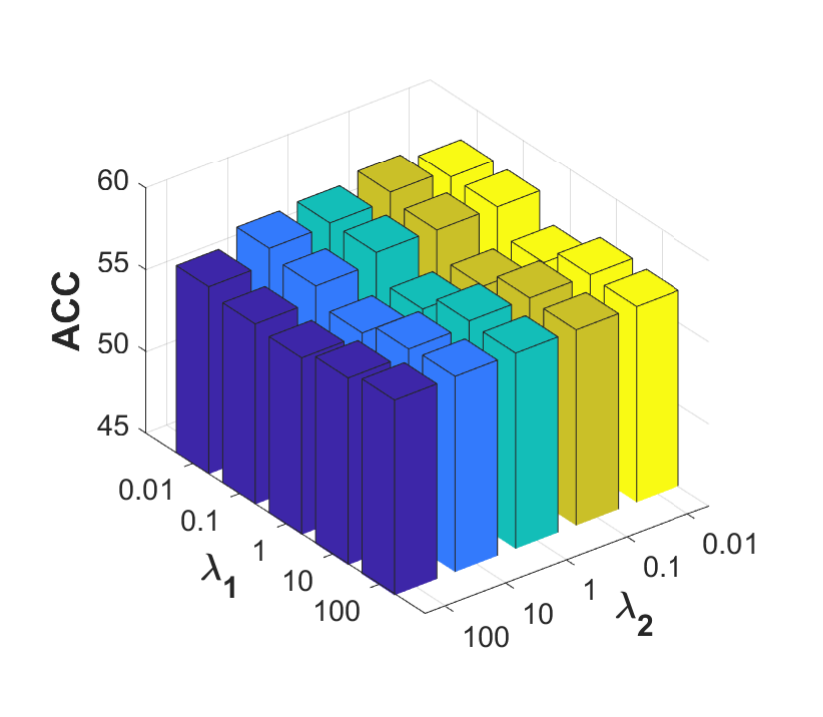}
	}
	\subfigure[UAT]{
		\includegraphics[width=0.31\linewidth]{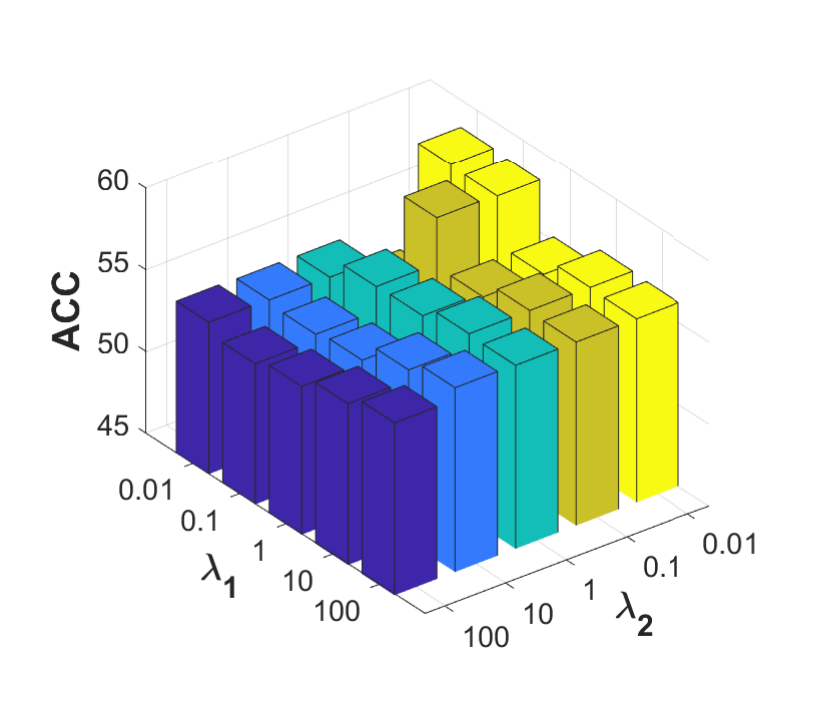}
	}
	\caption{The performance of SECL with different hyper-parameter $\lambda_1$ and $\lambda_2$ on six datasets.}
	\label{lambda}
\end{figure*}

\subsection{Experimental Results}
In this section, we conduct experiments on six datasets, selecting 12 benchmark algorithms for comparison. To reduce the influence of randomness and outliers, SECL is run 10 times across all datasets to obtain the average scores and the corresponding standard deviations (mean±std), for metrics such as ACC, NMI, ARI, and F1. For some baselines, we directly reference the results from \cite{liu2023simple,ding2023graph,wu2024deep}. Detailed experimental results are summarized in the Table \ref{EXP}, with the best results highlighted in bold and the runner-up results underlined. 

From the results presented in the table, it is evident that our proposed method achieves better performance in most cases, indicating that our approach is effective when compared to the benchmark algorithms being evaluated. Specifically, we make the following observations:
\begin{itemize}
	\item K-means and DEC methods did not achieve satisfactory results across all datasets. In contrast, methods based on Graph Neural Networks (GNNs) and contrastive learning generally outperformed K-means and DEC approaches. This suggests that leveraging both attribute information and structural information of the network is superior to using only one type of information. For instance, our method, when applied to the CORA dataset, saw improvements over DEC in ACC, NMI, ARI, and F1 score by about 60.84\%, 141.63\%, 249.24\% and 86.82\%, respectively.
	\item Our proposed method has achieved better clustering results compared to representative deep graph clustering methods. Taking the CORA dataset as an example, SECL surpassed GAE by about 72.41\%, 97.64\%, 221.61\%, and 118.91\% and DAEGC by about 6.19\%, 7.54\%, 6.47\%, and 7.35\% on ACC, NMI, F1, and ARI, respectively. This indicates that our contrastive learning strategy is capable of learning higher-quality node representations.
	\item The SCGC method demonstrates good performance relative to recent contrastive learning approaches, indicating that maintaining cross-view structural consistency is beneficial for learning discriminative node representations. Our method outperforms SCGC, showing even greater improvements on the CORA dataset with performance increases of 1.23\%, 1.39\%, 2.03\% and 3.50\% for ACC, NMI, ARI and F1, respectively. This may be due to the use of the modularity maximization and cross-view contrast loss.
	\item Our approach outperformed most of baseline algorithms. For instance, on the UAT dataset, it surpassed the runner-up method by about 3.01\%, 3.17\%, 2.38\% and 2.05\% for ACC, NMI, F1, and ARI, respectively. These results underscore the powerful capability of our SECL method in graph clustering, which can be attributed to our cross-view contrastive loss module, the structural contrastive loss module, and the modularity maximization module.
\end{itemize}
\begin{figure*}[!t]
	\centering
	\subfigure[CORA]{
		\includegraphics[width=0.26\linewidth]{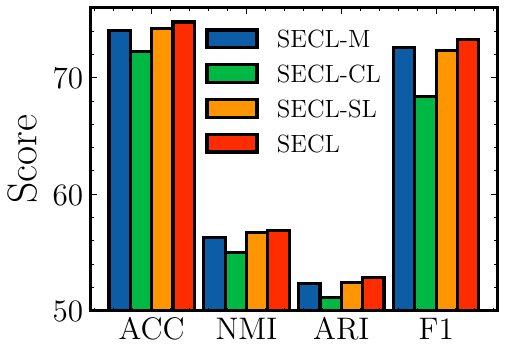}
	}\hspace{10mm}
	\subfigure[CITESEER]{
		\includegraphics[width=0.26\linewidth]{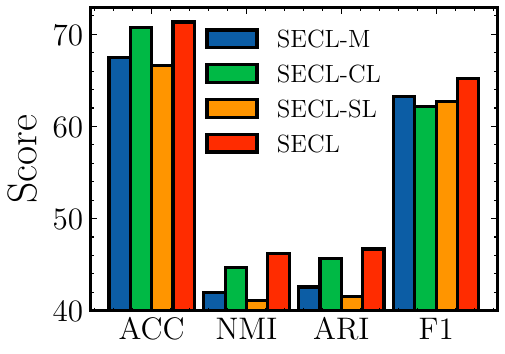}
	}\hspace{10mm}
	\subfigure[AMAP]{
		\includegraphics[width=0.26\linewidth]{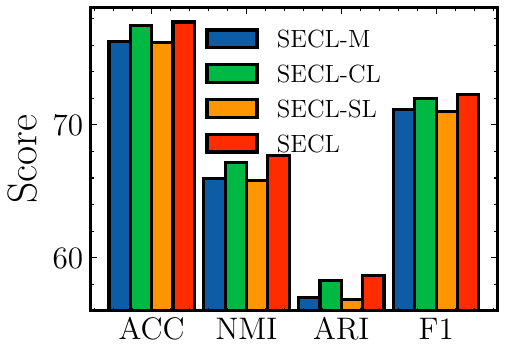}
	}
	\subfigure[BAT]{
		\includegraphics[width=0.26\linewidth]{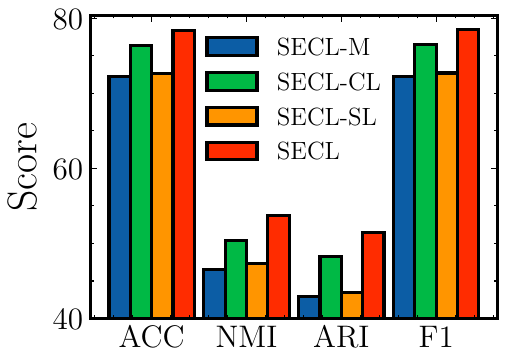}
	}\hspace{10mm}
	\subfigure[EAT]{
		\includegraphics[width=0.26\linewidth]{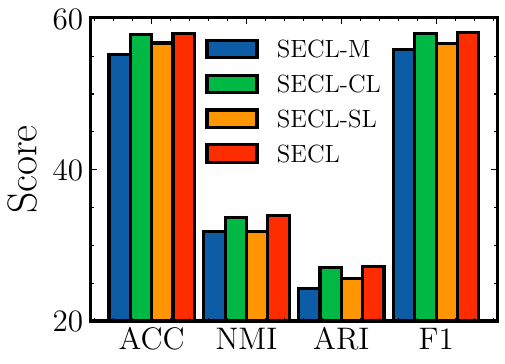}
	}\hspace{10mm}
	\subfigure[UAT]{
		\includegraphics[width=0.26\linewidth]{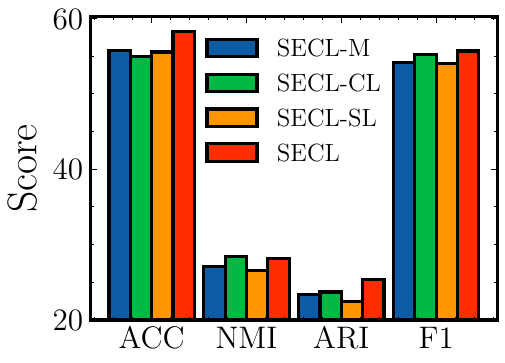}
	}
	\caption{Ablation comparisons of SECL on six datasets. (a), (b), (c), (d), (e) and (f) represent the results on CORA, CITESEER, AMAP, BAT, EAT and UAT, respectively.}
	\label{ab}
\end{figure*}
\subsection{Parameter Analysis}

\textbf{Hyper-parameter $\lambda_1$ and $\lambda_2$}: We conducted experiments to investigate the hyper-parameter $\lambda_1$ and $\lambda_2$, which balance the cross-view contrastive loss, structural contrastive loss and modularity maximization loss. For hyper-parameter $\lambda_1$ and $\lambda_2$, we selected from [0.01, 0.1, 1.0, 10, 100]. According to the experimental results shown in Figure \ref{lambda}. We can see that: The best result on the CORA dataset was obtained when $\lambda_1 =0.1$ and $\lambda_2 =0.01$. The best result on the AMAP dataset was obtained when $\lambda_1 =0.1$ and $\lambda_2 =10$. This indicates that balancing the three parts of the loss is important. 

\begin{figure}[h]
	\centering
	\includegraphics[width=0.99\columnwidth]{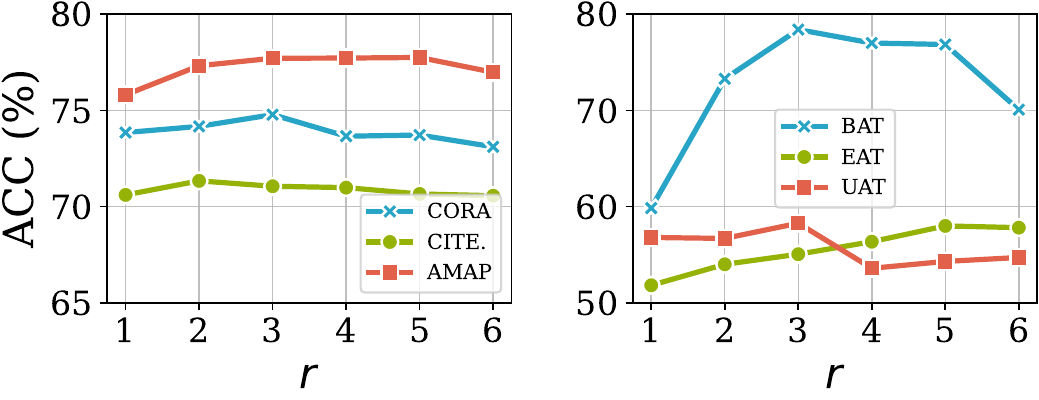}
	\caption{Sensitivity analysis of the number of graph Laplacian filters $r$.}
	\label{fig:r}
\end{figure}

\begin{figure}[h]
	\centering
	\includegraphics[width=0.99\columnwidth]{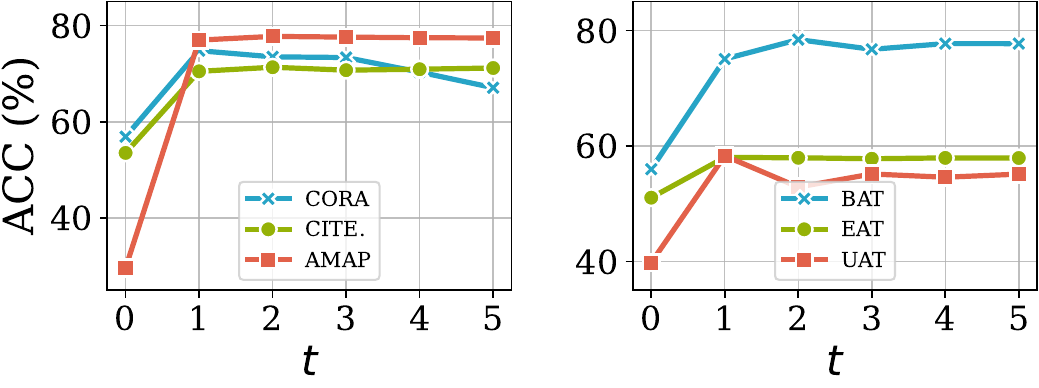}
	\caption{Sensitivity analysis of the number of $\mathrm{MLP}$ layers $t$.}
	\label{fig:t}
\end{figure}

\textbf{Sensitivity Analysis of hyper-parameters $r$ and $t$.}We conducted experiments to examine the effects of the number of layers $r$ in graph Laplacian filters and the number of layers $t$ in multi-layer perceptrons (MLPs). As shown in Figure~\ref{fig:r}, SECL demonstrates strong performance with $r$ values between 2 and 3. 
Moreover, our model becomes less sensitive to variations in $r$ for values in the range $3<r\leq 6$. In Figure~\ref{fig:t}, we observe that SECL achieves optimal performance with $t=2$ on the BAT, CITESEER, and AMPA datasets, while it performs best with $t=1$ on the other datasets. This indicates that our approach does not necessitate deep MLPs, effectively reducing the overall number of model parameters.

\subsection{Ablation Study}
In this section, ablation experiments are conducted to investigate the effectiveness of the SECL model. Figure \ref{ab}. displays the results of these experiments on all datasets. The comparison experiments are delineated as follows:
\begin{itemize}
	\item\textbf {SECL-M}: SECL without the modularity maximization module. The total loss of SECL-M canbe described by $\mathcal{L_{SL}+L_{CL}}$.
	\item\textbf {SECL-CL}: SECL without the cross-view contrastive module. The total loss of SECL-CL canbe described by $\mathcal{L_{SL}-L_{M}}$.
	\item\textbf {SECL-SL}: SECL without structural contrastive. The total loss of SECL-M canbe described by $\mathcal{L_{CL}-L_{M}}$.
\end{itemize}
Figure \ref{ab}. demonstrates the superior performance of our SECL model in all three ablation experiments, indicating the contribution of each module to the overall performance. For instance, the SECL enhances ACC, NMI, ARI, and F1 by approximately 5.66\%, 10.14\%, 9.70\%, and 3.13\%, respectively, in comparison to the SECL-M on CITESEER. This demonstrates the ability to attain higher-quality representations through collaborative optimization of modularity. Compared to the SECL-CL on CITESEER, the SECL enhances ACC, NMI, ARI, and F1 by approximately 0.82\%, 3.26\%, 2.17\%, and 4.91\%, respectively, demonstrating the necessity of the cross-view contrastive module. In comparison to the SECL-SL on CITESEER, the SECL significantly enhances ACC, NMI, ARI, and F1 by approximately 7.15\%, 12.22\%, 12.45\%, and 4.05\%, respectively, thereby emphasizing the effectiveness of the structural contrastive module.

\begin{figure*}[!t]
	\centering
	\subfigure[K-means]{
		\includegraphics[width=0.25\linewidth]{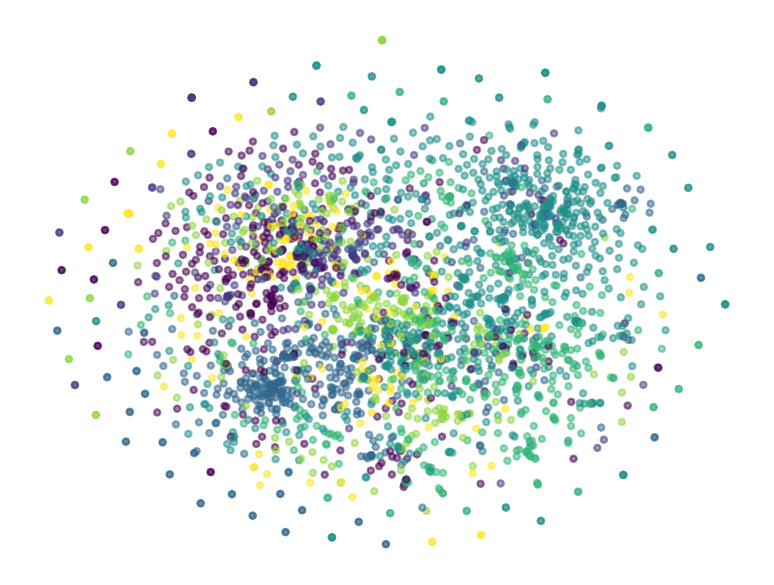}
	}\hspace{15mm}
	\subfigure[DEC]{
		\includegraphics[width=0.25\linewidth]{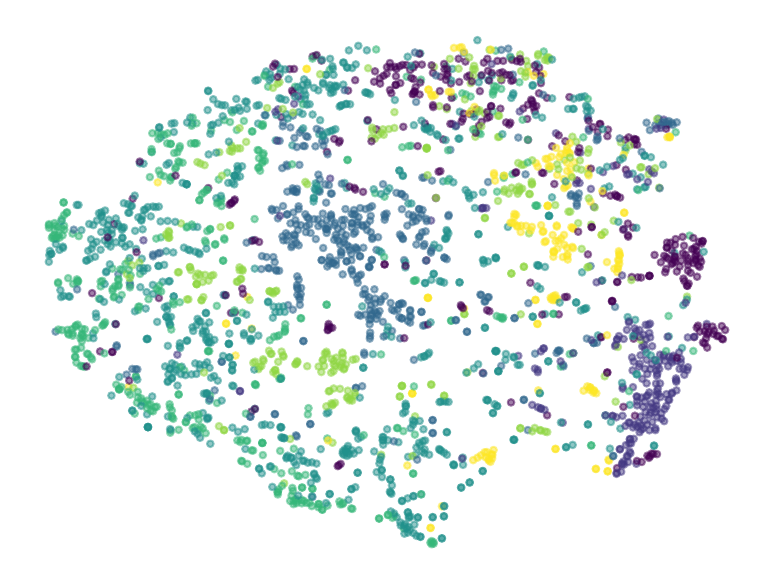}
	}\hspace{15mm}
	\subfigure[GAE]{
		\includegraphics[width=0.25\linewidth]{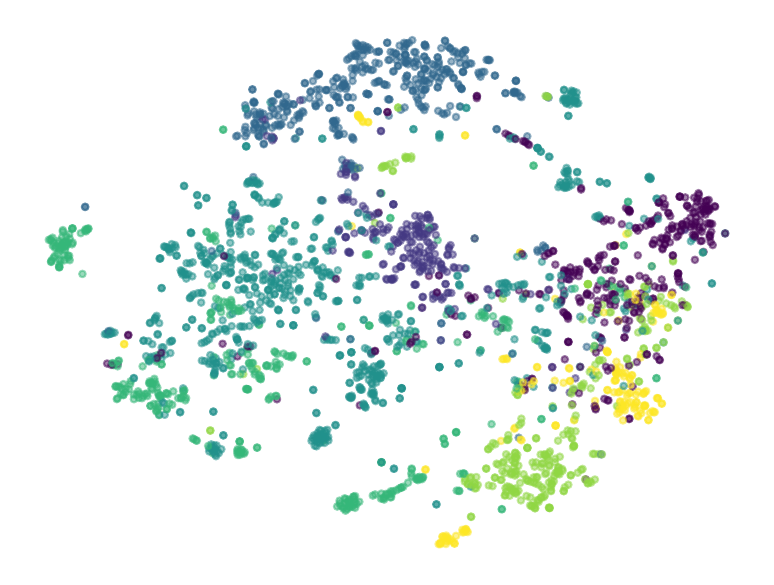}
	}\\
	\subfigure[DAEGC]{
		\includegraphics[width=0.25\linewidth]{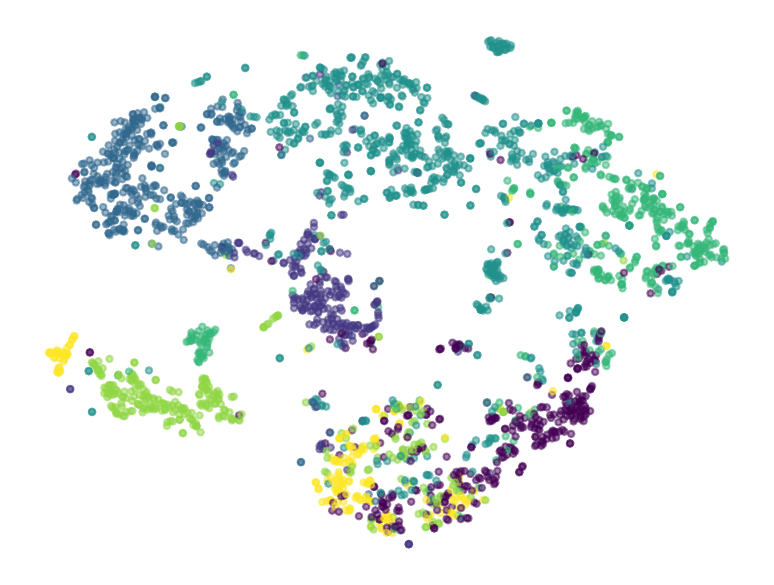}
	}\hspace{15mm}
	\subfigure[SCGC]{
		\includegraphics[width=0.25\linewidth]{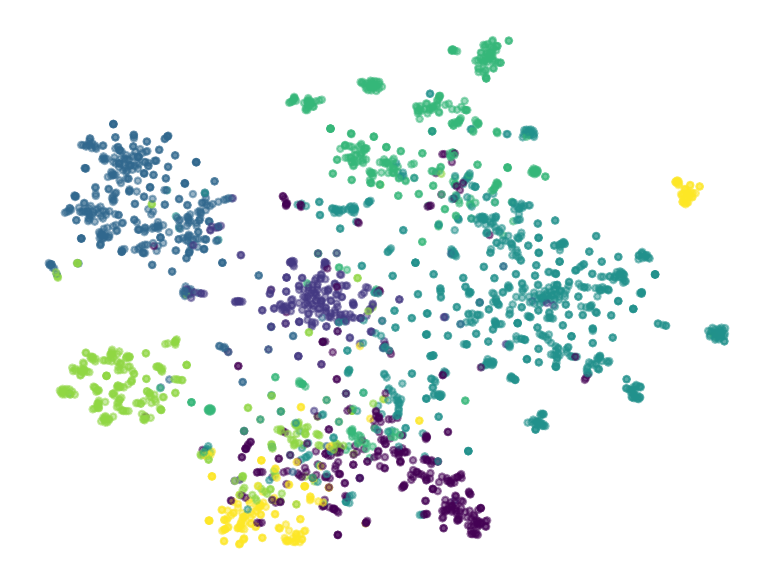}
	}\hspace{15mm}
	\subfigure[SECL]{
		\includegraphics[width=0.25\linewidth]{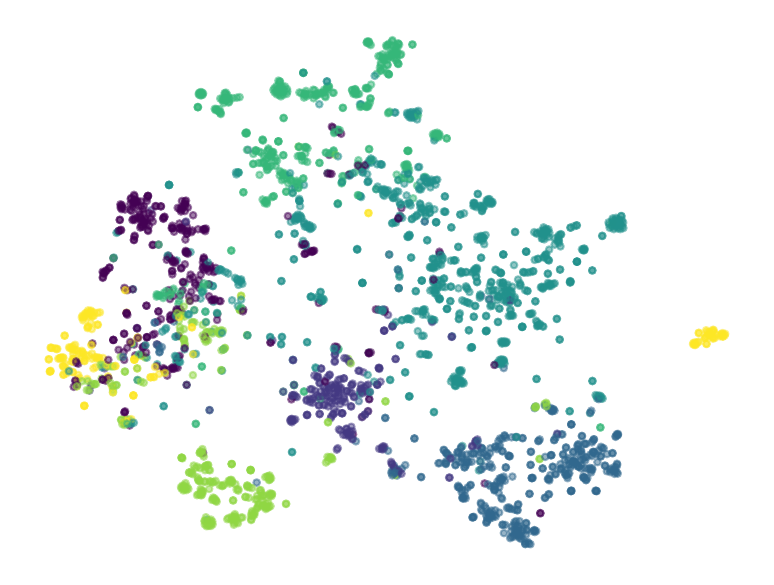}
	}
	\caption{Visualization of the SECL on CORA. (a), (b), (c), (d), (e) and (f) represent the visualization of CORA with K-means on raw features, DEC, GAE, DAEGC, SCGC and SECL, respectively.}
	\label{visualization}
\end{figure*}
\subsection{Visualization}
To demonstrate the superiority of our SECL method, we utilized the t-SNE \cite{van2008visualizing} (t-Distributed Stochastic Neighbor Embedding) technique to visualize the embeddings learned by the SECL method along with five other methods. Figure \ref{visualization}. displays the visualization results of these six methods on the CORA dataset. From the figure, it can be seen that our SECL method has achieved better performance compared to other methods, is able to reveal the underlying cluster structure more effectively. Compared to other algorithms, the clustering visualization results obtained from DAEGC display greater inter-cluster distances. This can be attributed to the utilization of KL divergence. Our results demonstrate tighter intra-cluster clustering and higher accuracy compared to DAEGC. This improvement can be attributed to our adoption of the modularity maximization module.

\begin{figure}[!t]
	\centering
	\subfigure[CORA]{
		\includegraphics[width=0.46\linewidth]{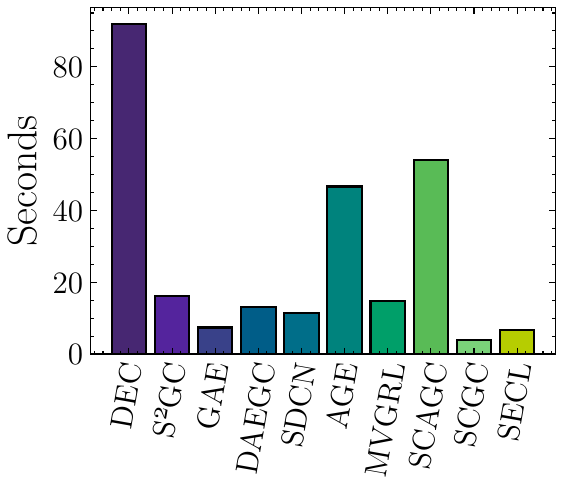}
	}
	\subfigure[CITESEER]{
		\includegraphics[width=0.47\linewidth]{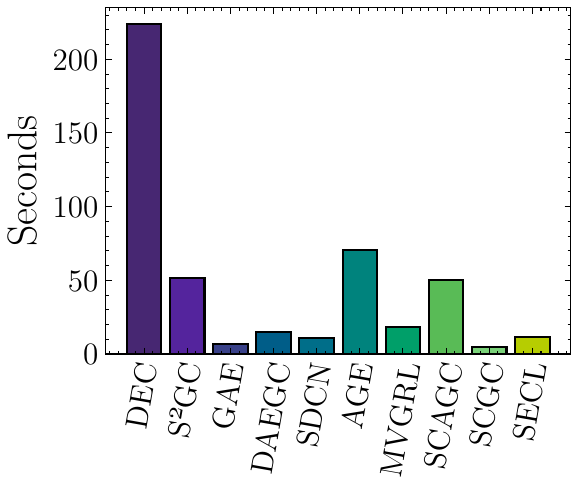}
	}
	\subfigure[CORA-DEC]{
		\includegraphics[width=0.46\linewidth]{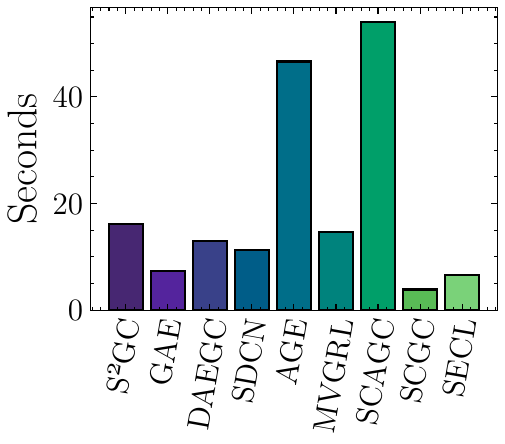}
	}
	\subfigure[CITESEER-DEC]{
		\includegraphics[width=0.46\linewidth]{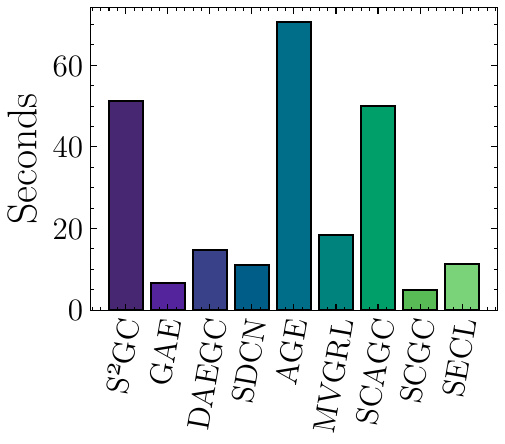}
	}
	\caption{Training time on the CORA and CITESEER datasets. (a) and (b) denote the training time on CORA and CITESEER. (c) and (d) represent the training time on CORA and CITESEER without DEC.}
	\label{time}
\end{figure}
\subsection{Time Costs}
In this section, the running times are reported for our SECL method and other algorithms on six datasets. The comparison includes the classic clustering algorithm DEC, deep graph clustering algorithms GAE, DAEGC, and contrastive learning methods such as AGE, MVGRL, SCAGC, and SCGC. To ensure a fair comparison, all algorithms were trained for 400 epochs. It is observed from Figure \ref{time}. (a) and (b) that DEC exhibits poor efficiency and achieves subpar results. To better showcase the running time, we have excluded the longest-running DEC, and the results are displayed in Figure \ref{time}. (c) and (d). Methods based on GCN and contrastive learning have improved efficiency, with SCGC being the most efficient. Our method is a little slower than SCGC in terms of efficiency. Although our algorithm has a slightly longer running time compared to SCGC, it demonstrates superior performance in experimental results, effectively balancing efficiency and effectiveness. Moreover, in comparison with other algorithms, our SECL method exhibits commendable performance in both running time and experimental results. 
\section{Conclusion}
This study has introduced SECL, a comprehensive framework for graph clustering that combines contrastive learning with modularity optimization. By utilizing inherent graph structures and attributes with two MLPs, SECL effectively obviates the need for pre-training and complex data augmentation. Then, a cross-view contrastive loss is proposed to enhance the discriminative capability of the node representations. Next, within the structure contrastive loss module, consistency of structure information is ensured by aligning the similarity matrix with the neighboring structure information. Finally, a modularity maximization module is employed to capture cluster-oriented information. Experimental results have confirmed the superiority of SECL over existing methods, demonstrating its ability to discern complex community structures within graphs. Our ablation studies have underscored the importance of each proposed module, while visualizations of the clustering results have offered intuitive evidence of the method's effectiveness. Despite a marginal increase in computational cost compared to the fastest baseline, the performance gains of SECL warrant its use. The success of SECL indicates a promising avenue for future research in graph clustering, with potential for extending to more complex networks and real-world applications. Hence, in future research, we aim to extend this method and develop new deep graph clustering models capable of handling large-scale and dynamic graph data.

\ifCLASSOPTIONcompsoc
  \section*{Acknowledgments}
\else
  \section*{Acknowledgment}
\fi
We would like to express our gratitude for the GitHub repository [A Unified Framework for Deep Attribute Graph Clustering] \footnote{https://github.com/Marigoldwu/A-Unified-Framework-for-Deep-Attribute-Graph-Clustering} and the GitHub repository [Awesome Deep Graph Clustering] \footnote{https://github.com/yueliu1999/Awesome-Deep-Graph-Clustering}. These repositories have significantly facilitated the implementation of our code and the conduct of our experiments. We deeply appreciate their valuable open-source contributions. This work is supported by the National Natural Science Foundation of China (Grant No. 61872432), the Shaanxi Fundamental Science Research Project for Mathematics and Physics (Grant No. 22JSY024).


\ifCLASSOPTIONcaptionsoff
  \newpage
\fi



%


\bibliographystyle{IEEEtran}
\bibliography{ref}
%
\begin{IEEEbiography}[{\includegraphics[width=1in,height=1.25in,clip,keepaspectratio]{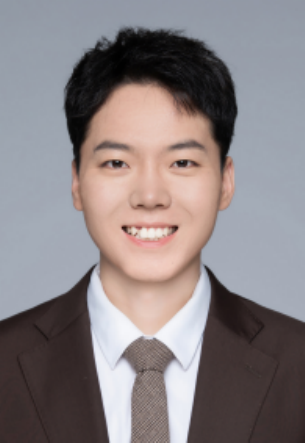}}]{Xunlian Wu}
received the B.S. degree in Electronical Information Science and Technology from ZhengZhou University, ZhengZhou, China in 2017. He is currently working toward the Ph.D degree in the school of Computer Science and Technology, Xidian University, Xi’an, China. His research interests include data mining and social network analysis.
\end{IEEEbiography}
\begin{IEEEbiography}[{\includegraphics[width=1in,height=1.25in,clip,keepaspectratio]{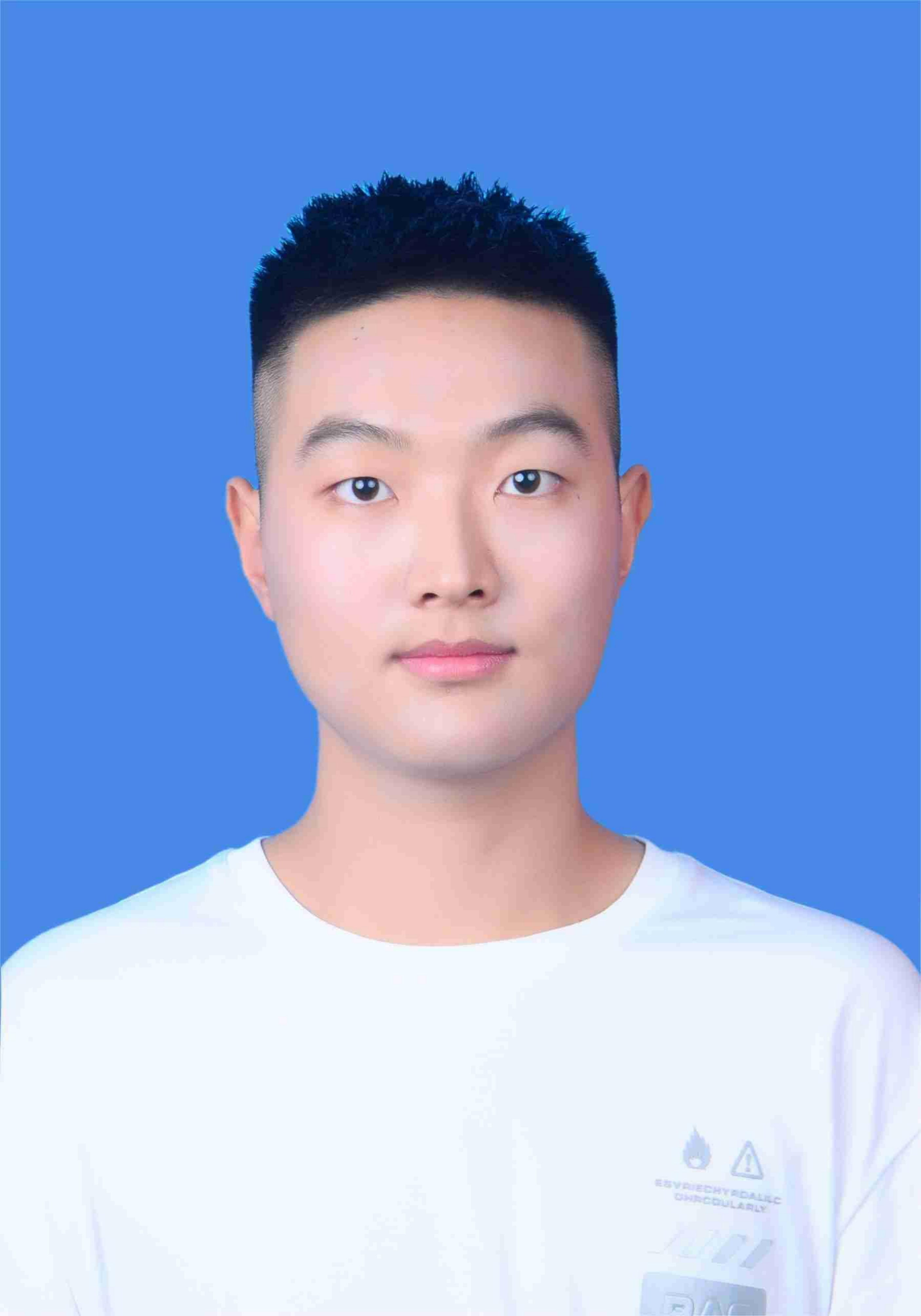}}]{Jingqi Hu}
JingQi Hu received the B.S. degree in Software Engineering from Xidian University, Xi'an, China in 2024. He is currently working toward the Master`s Degree in the school of Computer Science and Technology, Xidian University, Xi'an, China. His research interests include data mining and social network analysis.
\end{IEEEbiography}
\begin{IEEEbiography}[{\includegraphics[width=1in,height=1.25in,clip,keepaspectratio]{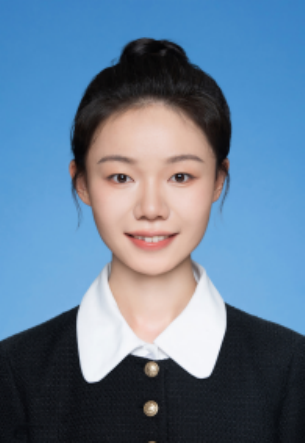}}]{Anqi Zhang}
is a lecture of School of Computer and Artificial Intelligence, Henan Finance University, Zhengzhou, 450046, China. She received the M.S. degree in Biomedical Engineering from Northwestern Polytechnical University, Xi’an, China. in 2020. Her research interests include medical data mining and graph clustering.
\end{IEEEbiography}
\begin{IEEEbiography}[{\includegraphics[width=1in,height=1.25in,clip,keepaspectratio]{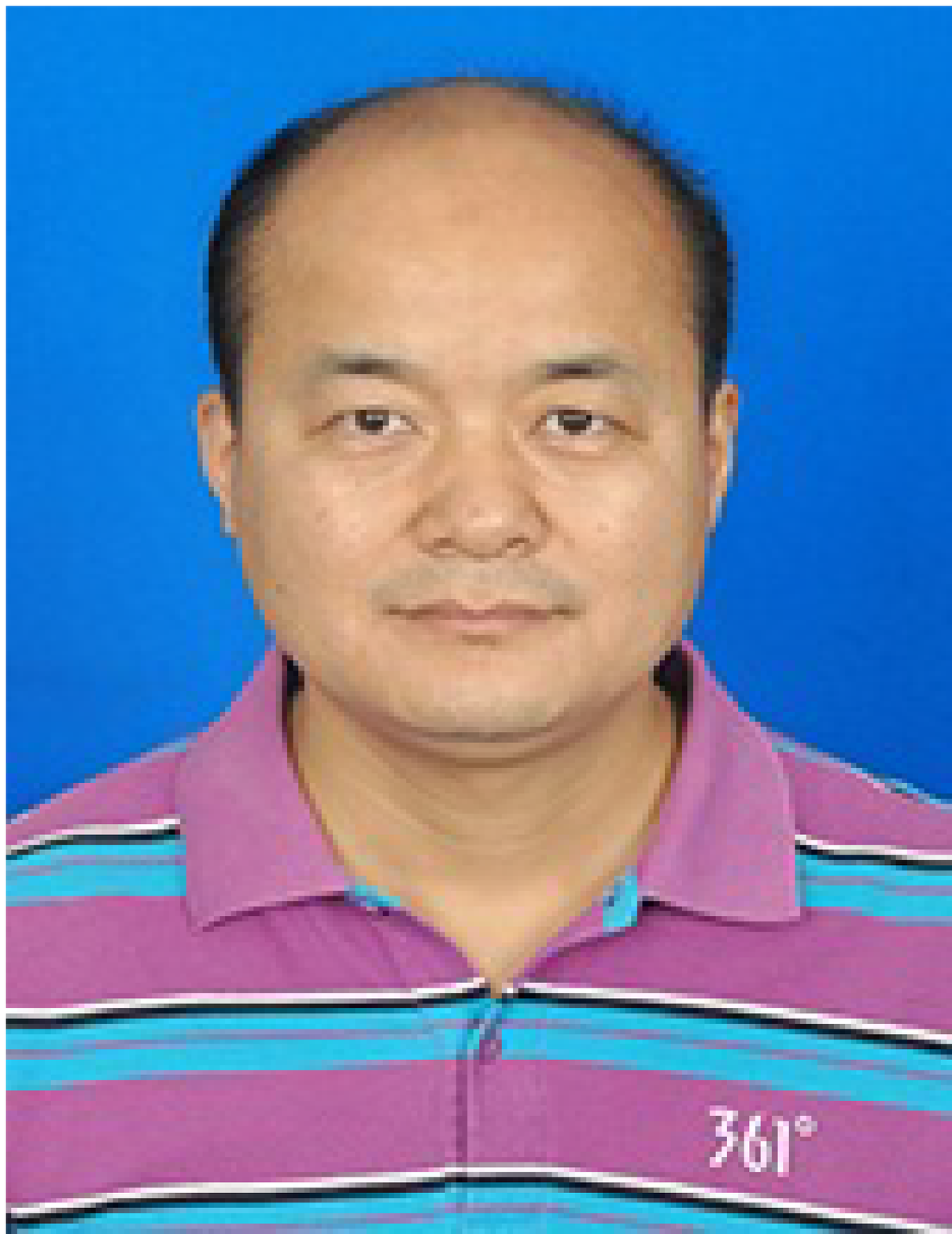}}]{Yining Quan}
is an associate professor at the School of Computer Science and Technology, Xidian University, Xi'an, 710071, China. His research interests include data mining, social network analysis, visualization and visual Analysis of big data.
\end{IEEEbiography}
\begin{IEEEbiography}[{\includegraphics[width=1in,height=1.25in,clip,keepaspectratio]{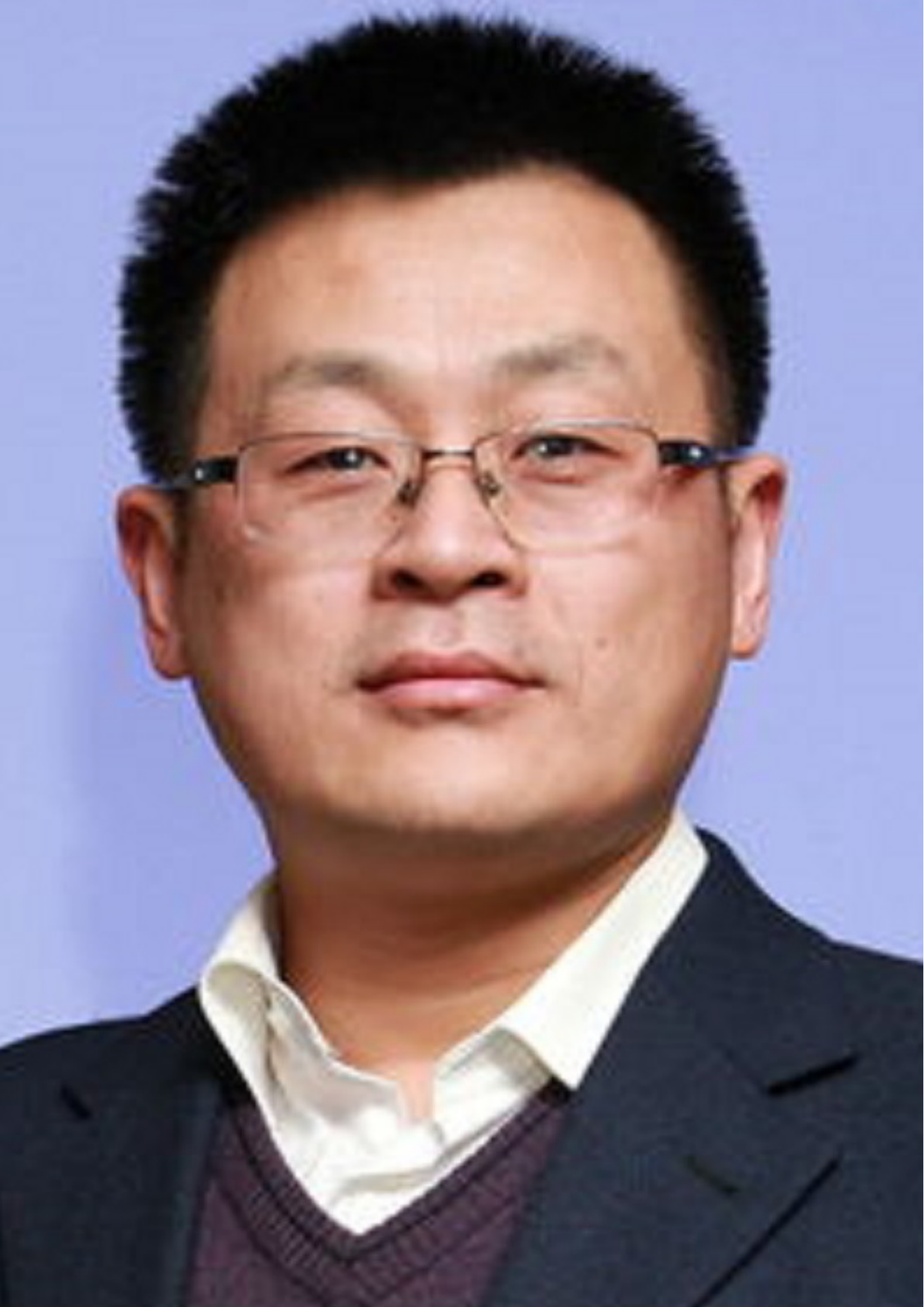}}]{Qiguang Miao}
is a professor and Ph.D. student supervisor of School of Computer Science and Technology in Xidian University. He received his doctor degree in computer application technology from Xidian University in December 2005. His research interests include machine learning, intelligent image processing and malware behavior analysis and understanding. In recent years, He has published over 100 papers in the significant domestic and international journals or conferences.
\end{IEEEbiography}
\begin{IEEEbiography}[{\includegraphics[width=1in,height=1.25in,clip,keepaspectratio]{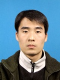}}]{Peng Gang Sun}
is an associate professor at the School of Computer Science and Technology, Xidian University, Xi'an, 710071, China. His research interests include data mining, pattern recognition, graph theory and complex network analysis.
\end{IEEEbiography}







\end{document}